\documentclass[12pt]{article}
\usepackage{amsmath,amsthm,amssymb,amsfonts}
\usepackage{algorithm}
\usepackage{algpseudocode}
\usepackage{mathtools}
\usepackage{graphicx}
\usepackage{tabularx}
\usepackage{xcolor}
\usepackage{hyperref}
\usepackage{caption}
\usepackage{subcaption}
\usepackage{float}
\usepackage{booktabs}
\usepackage{url}
\usepackage{microtype}
\usepackage{authblk}
\usepackage{enumitem}

\usepackage[margin=1in]{geometry}

\title{Evaluating Synthetic Tabular Data Generated To Augment Small Sample Datasets}

\author{Javier Mar\'in}
\affil{javier@jmarin.info}

\date{Last revision in March 15, 2025}
\theoremstyle{definition}
\newtheorem{definition}{Definition}[section]
\begin{document}

\maketitle

\begin{abstract}
This work proposes a method to evaluate synthetic tabular data generated to augment small sample datasets. While data augmentation techniques can increase sample counts for machine learning applications, traditional validation approaches fail when applied to extremely limited sample sizes. Our experiments across four datasets reveal significant inconsistencies between global metrics and topological measures, with statistical tests producing unreliable significance values due to insufficient sample sizes. We demonstrate that common metrics like propensity scoring and MMD often suggest similarity where fundamental topological differences exist. Our proposed normalized Bottleneck distance based metric provides complementary insights but suffers from high variability across experimental runs and occasional values exceeding theoretical bounds, showing inherent instability in topological approaches for very small datasets. These findings highlight the critical need for multi-faceted evaluation methodologies when validating synthetic data generated from limited samples, as no single metric reliably captures both distributional and structural similarity.

\vspace{0.5em}
\end{abstract}

\section{Tabular synthetic data}
Because of considerations such as privacy constraints, bias in some algorithms' training, and improving existing data quality, the use of generative models to build synthetic data is becoming more widespread. In this paper, we will address the specific problem of generating synthetic tabular data with a larger number of samples than the real data. The existing literature does not explicitly address this issue but instead refers to the generation and evaluation of synthetic data, presuming that increasing the quantity of samples is not the primary goal \cite{Benaim2020,Cheng2019,Eno2008,Goncalves2020,Hou2022,Karras2018,Rashid2019}.

Tabular data is one of the most prevalent data sources in data analysis. Commonly, this data is considered as "small data". This data is widely used in applications such as academics, business management, and any other case where a spreadsheet is used to collect data from multiple variables in tables. These tables have a small number of samples (rows) and frequently several columns.

The use of low sample data with machine learning algorithms is frequently questioned since the sample size is considered insufficient for producing reliable conclusions. Most machine learning applications involve fitting a probability distribution over a vector of observations. When the number of observations is small, algorithms tend to overfit to the training examples. In this context, the problem arises when users of these tables want to gain insights into the data and even be able to make predictions using machine learning.

Furthermore, most non-parametric tests used to validate these models are supported on the use of a sufficiently large number of observations. As a result, users of low sample tabular data do not have access to these technologies. For the experiments in this research, we have used the python open source project nbsynthetic that uses a non-conditional Wassertein-GAN that will be introduced in the next section. This library is specifically designed for low sample synthetic data generation and augmentation.

\subsection{Generative Adversarial Nets (GANs)}
Low sample data has the particularity of setting less constraints to generalization process in NN. Deep Neural Networks (DNN) are designed to ingest a large number of input samples with a large amount of training parameters in order to improve the accuracy of a given task \cite{Elman1993}. A common generalization method is the maximum likelihood estimation or MLE. MLE works by maximizing the probability of the observed data according to the model. When Goodfellow \cite{Goodfellow2014b} proposed the generative adversarial networks, his idea of regularization was not based on MLE. Instead, he proposed a method derived from noise-contrastive estimation or NSE. NSE uses an arbitrary, fixed "noise" distribution for the generator $p_g$ instead of the real input distribution. This is done by defining a parametric family of densities $p_\theta$ where the objective is to minimize the distance between the density $p_\theta$ and the input data distribution $p_r$ (that can be done by minimizing the Jensen-Shannon divergence), varying the parameter $\theta$. The goal GAN proposed is to learn a distribution that represents the source of synthetically generated data, where the distribution of a discriminator D, a binary classifier, $p_c$ that tries to figure out whether a particular sample was selected from training data or from a "generator" distribution, $p_g$, is parameterized directly.

GANs minimize a convex divergence within a parametric density space, which ensures asymptotic consistency \cite{Goodfellow2014}. When the model has sufficient capacity, it reaches a unique global optimum where the model distribution accurately matches the real data distribution.

The Wassertein GAN we are going to use in this paper is based on the fact that Wassertein distance or Earth mover distance, $W(p_r, p_\theta)$ may have better qualities on comparing probability distributions than, for example, the Jensen-Shannon divergence \cite{Arjovsky2017}.

\subsection{The GAN convergence problem}
The question of training GAN with low sample data has been already studied by several authors \cite{Chaudhari2020,Liu2019}. Even if we want to increase the amount of synthetic samples compared to the original data, generating synthetic data from original samples is a relatively simple task. GANs, like other NN, struggle to capture the underlying structure of the data with low sample training data, and actually with fewer samples than dimensions. The discriminator attempts to identify samples from the training data or generator distribution, and while it initially converges, it may drift apart due to overfitting if there is not enough data. The generator may also be unable to reach a stable state with limited data, resulting in a lack of convergence and inadequate weight/bias adjustments. Furthermore, it may be difficult to split the data into training, test, and validation sets when dealing with small datasets, making it difficult to assess the model bias.

A key question is: How many samples are needed for effective GAN augmentation? While neural networks typically require about 10 samples per dimension, GANs operate differently. Unlike maximum likelihood estimation methods bound by the Central Limit Theorem, GANs use adversarial learning where two networks compete. This process converges quickly to a local Nash equilibrium at a linear rate \cite{Mazumdar2020}, potentially requiring fewer samples. However, this doesn't make GAN training simpler—it remains more complex than traditional MLE approaches.

Figure 1 compares WGAN training on two datasets: small (9×8) and large (1055×42). With the small dataset, the generator's learning plateaus after about 15 steps, and the discriminator becomes unstable and fails to converge. With the large dataset, the discriminator converges successfully throughout training. For small datasets, the generator quickly learns an approximation of the underlying distribution—one that our experiments show is surprisingly accurate—but the discriminator eventually begins classifying inputs randomly, a behavior predicted by Goodfellow \cite{Goodfellow2014}.

\begin{figure}[h]
\centering
\includegraphics[width=0.75\textwidth]{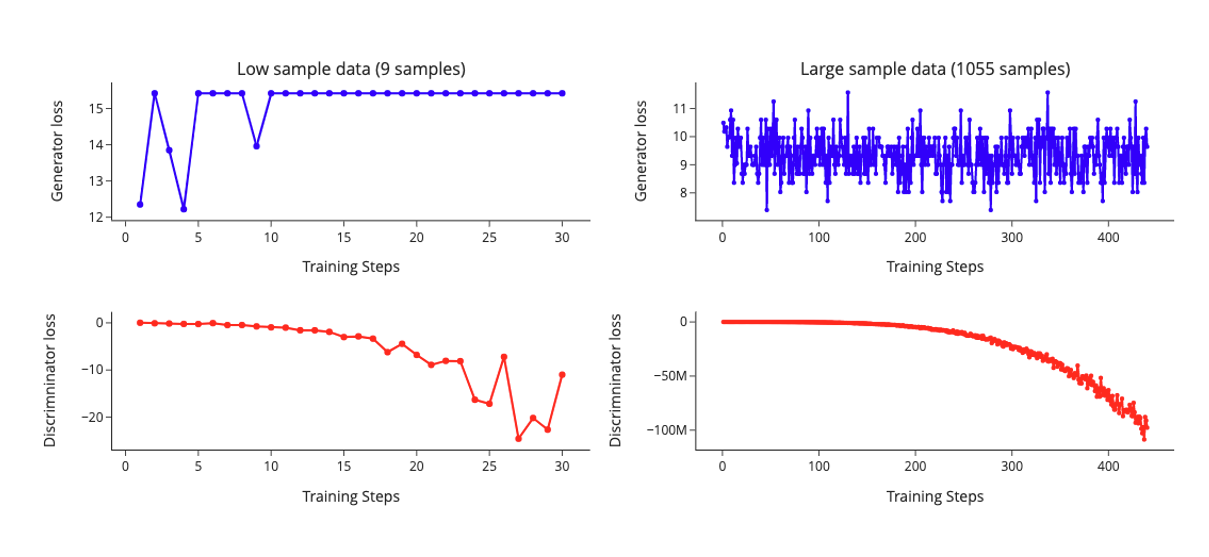}
\caption{Model convergence comparison using a WGAN for low and large sample training data. Wassertein GAN model trained with batch size $M = 10$, $n_{critic} = 1$, clipping parameter, $c = 0.01$, $\alpha = 10^{-5}$, and RMSprop optimizer. Left plot: datasets with 9 samples. Right plot: dataset with 1055 samples.}
\label{fig:model-convergence}
\end{figure}

As we already pointed, despite this instability, generator $G$ will learn an approximate distribution from training data, and discriminator will learn to predict data class quite accurately. Also, if we perform an univariate test for comparing original and synthetic distributions, we will observe that both are drawn from the same probability distribution. But we think that this is not enough. The key will be to validate if patterns in this synthetic generated data will be useful for using this data with other machine learning algorithms. The main goal of our research is to figure out how to do it.

\section{Global data metrics}
In this research, we make a critical examination of relevant global data metrics identified in the literature \cite{Benaim2020,Borgwardt2006,Chundawat2022,Eno2008,Goncalves2020,Goodfellow2014,Rosenbaum1983,Snoke2018,Snoke2018a,Sutherland2016,Sutherland2017,Theis2016,Woo2009,Xu2018}. We deliberately exclude univariate and bivariate statistical tests from our analysis, as they present fundamental limitations for synthetic data validation. It is important to acknowledge that establishing distributional equivalence at the individual variable level—while informative—is insufficient to prove functional equivalence of synthetic datasets. The verification that marginal or pairwise distributions match between original and synthetic data does not guarantee that the complex interdependencies necessary for predictive modeling are preserved. This distinction is particularly crucial when evaluating whether synthetic data can reasonably substitute original data in machine learning applications where multivariate relationships determine model performance. 

Global metrics offer a more comprehensive evaluation framework than univariate approaches by capturing not only the marginal distributions of individual variables but also their complex interdependencies. This section examines three established global metrics that have gained prominence in synthetic data validation: propensity score measurement (pMSE), Maximum Mean Discrepancy (MMD), and cluster analysis measures.
\begin{definition}
Let $X$ be a sub-space of $\mathbb{R}^d$ with metric $k$. Let $\text{Prob}(X)$
denote the space of probability measures defined on $X$. We define $p_o$ as the probability distribution of original data and $p_s$ as the distribution of synthetic data, where po,$p_o, p_s \in \text{Prob}(X)$
\end{definition}

These global metrics diverge methodologically from elementary distance calculations between distributions $p_o$ and $p_s$. Classical distances such as Total Variation and Kullback-Leibler divergence operate directly on distributional differences: TV measures the maximum probability difference between distributions, creating a metric space with distance $d_T$, while KL divergence fails to qualify as a true metric due to its asymmetry and violation of the triangle inequality. Instead, the global metrics we examine— Propensity Score Metric or pMSE, Maximum Mean Discrepancy or MMD, and cluster analysis measures—evaluate data similarity through distinct mechanisms focused on distributional characteristics, functional properties in reproducing kernel Hilbert spaces, and geometric clustering properties, respectively.

\subsection{Propensity score}
We selected this metric for our benchmark because propensity scores -pMSE- have become standard for data quality assessment \cite{Woo2009}. Unlike formal distances or divergences, they measure the conditional probability of treatment assignment given observed covariates \cite{Rosenbaum1983}. Samples with similar covariate distributions tend to have matching propensity score distributions, for both univariate and multivariate cases. To calculate propensity scores, we combine original and synthetic datasets, assigning the value $1$ to synthetic samples and $0$ to original samples. We then use a classification algorithm to compute each case's probability and compare the resulting propensity score distributions between datasets.

To calculate the propensity score, we can use a classification algorithm like logistic regression, where the predicted probability for each instance will be the values of the propensity score. \cite{Snoke2018} proposed the following definition:
$$pMSE = \frac{1}{N}\sum_{i=1}^{N}(\hat{p}_i - 0.5)^2$$
where $N$ is the number of samples, $\hat{p}_i$ the predicted probabilities, and pMSE is the mean-squared error of the logistic regression-predicted probabilities.

\subsection{Cluster analysis measure}
Cluster analysis measure use supervised learning to assess whether data instances belong to original or synthetic samples, similar to a GAN discriminator \cite{Goodfellow2014}. While GANs maximize correct label assignment probabilities, cluster analysis \cite{Woo2009} uses distance metrics like Euclidean distance to identify similarities. However, this approach becomes problematic in high-dimensional spaces due to the concentration of distances phenomenon \cite{Beyer1998}. In our research, this problem is not occurring. 

If we divide a dataset $X$ in two parts of size $\mathcal{L}$ and $\mathcal{L}'$, the percentage of observations in $\mathcal{L}$ is given by 

$$c = \frac{n_\mathcal{L}}{n_\mathcal{L} + n_{\mathcal{L}'}}$$

To proceed, we also combine the original and synthetic data sets, assigning a variable with a value of one to all synthetic data instances and a value of zero to all original data instances. Then, we perform a cluster analysis choosing a fixed number of clusters $C$ and calculate the following:

$$U_c = \frac{1}{C}\sum_{i=1}^{C}w_i\left[\frac{n_{io}}{n_i} - c\right]^2$$

where $n_i$ is the number of observations in the $i$-th cluster, $n_{io}$ is the number of observations from the original data in the $i$-th cluster, $w_i$ is the weight assigned to $i$-th cluster. High $U_c$ means that points in the original and synthetic data spaces are too separated to fall in the same cluster, suggesting different probability distributions $p_o, p_s$. Since both methods use supervised learning to identify sample origins rather than hypotheses testing, they basically compute class membership likelihoods.

While our previous exposition highlights significant limitations of distance-based metrics in high-dimensional spaces, we selected this metric despite its reliance on distance calculations for several complementary reasons.

First, our experimental context involving extremely low-sample datasets mitigates many of the typical dimensionality concerns. With sample sizes substantially smaller than dimensionality, the theoretical issues of distance concentration are overshadowed by the more immediate challenge of statistical power. In this specific context, the cluster analysis measure provides valuable geometric insights that complement our topological approaches.

Second, the cluster analysis measure offers an intuitive interpretative framework that bridges traditional statistical metrics and modern discriminative approaches. By directly assessing whether synthetic and original samples naturally fall into the same clusters, this metric provides a geometric perspective on data similarity that is readily comprehensible even to researchers without relevant background in topology or kernel methods.

Third, we deliberately include this potentially problematic metric to empirically demonstrate the inconsistencies between different evaluation approaches. By contrasting the results of distance-based clustering with our proposed topological methods across multiple datasets, we can concretely illustrate how different methodological assumptions lead to divergent conclusions about synthetic data quality.

Last, the cluster analysis measure is an established reference point in the literature, enabling meaningful comparison with previous research. 

\subsection{Maximum Mean Discrepancy}
The use of test statistics based on quantities defined in reproducing kernel Hilbert spaces ($RKHSs$) is an approach that has gained popularity over the past few years. Maximum Mean Discrepancy (MMD) is a non-parametric hypothesis testing statistic for comparing samples from two probability distributions \cite{Gretton2006,Gretton2009}. MMD uses a kernel to define the similarities between probability distributions. One of the most common kernels is Gaussian $rbf$. The function $rbf$ kernels computes the radial basis function (RBF) kernel between two vectors. This kernel is defined as: 
$$k(x, x') = \exp(-\gamma \|x - x'\|^2)$$
Small $\gamma$ values define a Gaussian function with a large variance. The advantage of this method is the use of a kernel that defines similarities between observations. Maximum Mean Discrepancy is defined as follows:
$$MMD(p, q) \coloneqq \|\mu_p - \mu_q\|^2$$
Maximum Mean Discrepancy (MMD) follows the fundamental distance properties articulated in Section 2, satisfying both non-negativity $MMD(p, q) \geq 0$ and uniqueness conditions ($MMD(p,q)=0$ if and only if $p = q$). These properties establish MMD as a valid measure of distributional dissimilarity, where values approaching zero indicate increasing similarity between distributions $p$ and $q$. However, unlike normalized metrics constrained to specific intervals, MMD lacks an upper bound, which presents interpretative challenges when comparing results across different datasets or problem domains. This unboundedness means that while two MMD values of precisely zero indicate identical distributional matching, comparing non-zero MMD values between different datasets requires careful contextual consideration, as the scale of MMD depends on the specific kernel chosen and the underlying data characteristics. This limitation becomes particularly relevant in our cross-dataset comparative analysis, where absolute MMD values cannot be directly interpreted as relative measures of synthetic data quality without appropriate normalization. 

\section{Topological data analysis}
Topological data analysis or TDA is a collection of methods that provides qualitative data analysis \cite{Carlsson2009}. The mathematical field of topology is concerned with the study of qualitative geometric information. In order to gain knowledge from high-dimensional data sets, the connectivity of the data, or how a data space's connected elements are organized, TDA has become an essential tool. The benefits of using data connectivity were emphasized by \cite{Carlsson2009} with the following statement:

\begin{quote}
"Topology studies geometric properties in a way which is much less sensitive to the actual choice of metrics than straightforward geometric methods, which involve sensitive geometric properties such as curvature \cite{Carlsson2009}"
\end{quote}

This insight has profound implications for small datasets. While high-dimensional limitations like distance concentration may not directly apply to our low-sample scenarios, the arbitrary nature of distance selection becomes even more consequential. With few observations, distance measures become highly sensitive to outliers, and the statistical power to validate distributional assumptions is very limited. Small perturbations in distance calculations can lead to substantially different conclusions about data similarity, a vulnerability increased by the discretization effects introduced by discrete variables in tabular data.
Moreover, as Carlsson emphasizes \cite{Carlsson2009}, conventional metrics impose rigid notions of proximity that fail to capture important topological features preserved under continuous deformations. In small datasets, these topological characteristics may be more informative than raw distances. The fundamental limitation is not computational complexity but interpretative validity—distance measures on small samples lack statistical robustness regardless of dimensionality. This critique needs approaches that observe data relationships through connectivity rather than arbitrary distance constructs, particularly when sample sizes preclude reliable distribution estimation.

\subsection{Homology and Persistence diagrams}
Homology is an algebraic invariant that counts the topological attributes of a space \cite{Carlsson2004}. The collection of these components is a qualitative invariant of the space. Spaces are path-connected in a specific way, and this property is preserved under continuous deformations. We summarize the topological information about the data space in the notion of a persistence module. Persistence homology is a mathematical formalism which allows to infer topological information from a sample of a geometric object. We can compare two data spaces with a mathematical formulation of these properties as line-connections (level-zero connectivity information), loops(level-one connectivity information) and so on.
\begin{definition}
For any topological space $X$, abelian group $A$, and integer $k \geq 0$, there is assigned a group $H_k(X, A)$.
\end{definition}

\begin{definition}
Filtered complex: Let $X$ denote a metric space, with metric $d$. Then the Vietoris-Rips complex for $X$, attached to the parameter $\epsilon$, denoted by $VR(X, \epsilon)$, will be the simplicial complex whose vertex set is $X$, and where $v_0, v_1, \ldots, v_k$ spans a $k$-simplex if and only if $d(v_i, v_j) \leq \epsilon$ for all $0 \leq i,j \leq k$.
\end{definition}
A simplicical complex is an expression of the space as a union of points, intervals, triangles, and higher dimensional analogues \cite{Carlsson2009}. Topological spaces can be approximated using simplicical complexes.

\begin{figure}[h]
\centering
\includegraphics[width=0.65\textwidth]{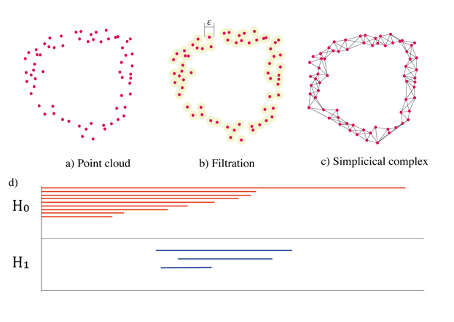}
\caption{The Rips complex on a point cloud $VR(X, \epsilon)$. a) A set of point cloud data $X$ b) Vietoris-Rips complex for $X$, $VR(X, \epsilon)$ c) $\mathbb{R}^2$ persistence simplicical complex as union of points, intervals, triangles, and higher dimensional analogues d) Betti numbers $\beta_0$ and $\beta_1$ for $H_0$ and $H_1$.}
\label{fig:rips-complex}
\end{figure}

\begin{definition}
Betti numbers: For any field $F$, $H_k(X, F)$ will be a vector space over $F$. Its dimension, if it is finite dimensional, will be written as $\beta_k(X, F)$, and will be referred to as the $k$-th Betti number with coefficients in F. The $k$-th Betti number corresponds to an informal notion of the number of independent $k$-dimensional surfaces. If two spaces are homotopy equivalent, then all their Betti numbers are equal \cite{Carlsson2009}.
\end{definition}

\begin{definition}
The $k$-th persistent homology of the filtration $X$ is the sequence
$$PH_k(X, F): H_k(X_0, F) \rightarrow H_k(X_1, F) \rightarrow H_k(X_k, F)$$
of vector spaces and induced linear transformations \cite{Barnes2021}.
\end{definition}
If each $H_k(X_i, F)$ is finite dimensional and $\beta_k^{i,\ell}(X_k, F)$ denotes the rank of the linear transformation $H_k(X_i, F) \rightarrow H_k(X_\ell, F)$ induced by the inclusion $X_i \subset X_\ell$, $i \leq \ell$, the persistence diagram of $PH_k(X, F)$ denoted $dgmn(X, F)$, is the collection of pairs $(i, \ell)$ with nonzero multiplicity.

\begin{definition}
A barcode is a finite set of intervals that are bounded below \cite{Carlsson2004}. Barcodes represent the space at various scales. They are a family of intervals with endpoints in $\mathbb{R}^+$ derived from homology vector spaces $\{H_i(VR(X, \epsilon))\}_{i \geq 0}$. Each such interval can be considered as a point in the set $D = \{(x,y) \in \mathbb{R}^2| x \leq y\}$.
Let $Y$ denote the collection of all possible barcodes. We need to define a quasi-metric $D(S_1, S_2)$ on all pairs of barcodes $(S_1, S_2)$, with $(S_1, S_2) \in Y$.
\end{definition}

\subsection{Stability of persistence diagrams}
Persistence diagrams are stable with respect to perturbations of the input filtration. This is known as the stability theorem \cite{Cohen-Steiner2005}.

\subsubsection{Distances between persistence diagrams}
To define a metric space over persistent diagrams, we require an appropriate distance function. Let $X$ be a topological space and $d : X \times X \rightarrow [0, \infty]$ such that:
\begin{itemize}
\item for all $x \in X$, $d(x, x) = 0$, - uniqueness
\item for all $x, y \in X$, $d(x, y) = (y, x)$, 
\item for all $x, y, z \in X$, $d(x, z) \leq d(x, y) + d(y, z)$ - triangle inequality
\end{itemize}
\begin{definition}
For any topological space $X$ be a sub-space of $\mathbb{R}^m$, we represent a metric space as $(X, d)$ for each $d : X \times X \longrightarrow \mathbb{R}^+$.
\end{definition}
\begin{definition}
The Hausdorff distance and the Bottleneck distance between $X$ and $Y$ are defined as
$$d_H(X, Y) = \max\{\sup_x \inf_y \|x - y\|_\infty, \sup_y \inf_x \|y - x\|_\infty\}$$

The bottleneck distance is based on a bijection between the points in a diagram.
$$d_B(X, Y) = \inf_\gamma \sup_x\|x - \gamma(x)\|_\infty$$

Distances $d_H$ and $d_B$ satisfy the triangle inequality and therefore are distances:
\begin{align*}
X = Y \text{ then } d_B = d_H = 0, \text{ and}\\
X \subseteq Y \text{ then } d_B = d_H = 0
\end{align*}

As a conclusion from the stability theorem , if two spaces or sets $X, Y \subset \mathbb{R}^n$ are Hausdorff close, then their barcodes are Bottleneck close. Given two persistence diagrams $X, Y$
$$\text{if } d_H(X, Y) < \epsilon, \text{ then } d_B(X, Y) < \epsilon$$

Since the bottleneck distance satisfies one more constraint, it is bounded below by the Hausdorff distance \cite{Cohen-Steiner2005}:
$$d_H(X, Y) \leq d_B(X, Y)$$
\end{definition}

\begin{definition}
A continuous function between two topological spaces $X$ and Y, abelian groups $X, Y$, induces linear maps between the homology groups. For an integer $k \geq 0$, we have
$$f_k: H_k(X) \rightarrow H_k(Y)$$
where $H_k$ denotes the $k$th singular homology group.
\end{definition}

\begin{definition}
Bottleneck stability: Let $X$ be a triangulable space with continuous tame functions, $f, g: X \rightarrow \mathbb{R}$. Then the persistence diagram satisfies $d_B(D(f),D(g)) \leq \|f - g\|_\infty$.
\end{definition}

This definition shows that persistence diagrams are stable under some perturbations of small amplitude. Intuitively, Bottleneck distance between two persistence diagrams can be seen as a measure of the cost of finding a correspondence between their points. High distances would imply that this cost is also higher, meaning that both diagrams don't match.

\subsubsection{Empirical distribution tests on barcodes}
While the homology of $X$ captures information about the global topology of the metric space, the probability space structure plays no role at all \cite{Blumberg2014}. This leads us to a fundamental limitation when we want to use hypothesis testing and confidence intervals to compare different topological invariants. We want to determine if we can reject the hypothesis that two empirical distributions on barcode space $Y$ (definition 3.1.5) came from the same underlying distribution. As the main asymptotic results on non-parametric tests for distribution comparison work only for distributions on $\mathbb{R}$, therefore the first transformation will be to map data from barcode space $Y \rightarrow \mathbb{R}$.

\begin{definition}
\cite{Blumberg2014} Let $(X, d_X, \mu_X)$ be a compact metric measure space and $k, n \in \mathbb{N}$. Let $\{x_1, x_2 \ldots, x_i\}$ a sequence of randomly drawn samples from $(X, d_X, \mu_X)$. $X_i$ is a metric measure space using the subspace metric and the empirical measure. Then $\Phi_k^n(X_i)$ converges in probability to $\Phi_k^n(X, d_X, \mu_X)$. For any $\epsilon > 0$
$$\lim_{n\to\infty} P \left| \Phi_k^n(X_i) - \Phi_k^n(X, d_X, \mu_X) > \epsilon \right| = 0$$
\end{definition}

\begin{definition}
$Y$ denote the collection of all possible barcodes. We can define a distance distribution $D$ on $\mathbb{R}$ to be the distribution on $\mathbb{R}$ induced by applying $d_L(-,-)$ to pairs $(S_1, S_2)$, drawn from $\Phi_k^n(X, d_X, \mu_X)^{\otimes 2}$ \cite{Blumberg2014}.
\end{definition}

\begin{definition}
Let $S$ be a barcode, $S \in Y$. $D_S$ will be the distribution induced by applying $d_L(S, -)$. Since $D$ and $D_S$ are continuous with respect to the Gromov-Prohorovmetric, $D_S$ converges in probability to $D$ \cite{Blumberg2014}.
\end{definition}

Therefore we can directly compare distributions on barcode spaces $D \in \mathbb{R}$ using hypothesis testing. One application of these projections is the use of the two-sample Kolmogorov-Smirnov statistic \cite{Massey1951}. This test statistic offers a method of assessing whether two empirical distributions observed were drawn from the same underlying distribution. The most important strength of this statistic is that, for distributions on $\mathbb{R}$, the test statistic's p-values are asymptotically independent of the underlying distribution provided. An alternative to the Kolmogorov-Smirnov test is the Mann-Whitney U test, $U$,  also known as the Wilcoxon rank sum test, and is commonly used to find differences between two groups on a single, ordinal variable with no specific distribution. Last, another suitable method for testing the independence of two barcodes distributions could be the Chi-square test of independence of variables in a contingency table \cite{Lin2015,Virtanen2020}.

\subsubsection{Normalized Bottleneck distance}
The instability of persistence diagrams and the unbounded nature of the Bottleneck distance make it impractical for measuring similarity between input data and synthetic data. Like spatial distances and statistical divergences (e.g., Kullback–Leibler and Jensen-Shannon divergences, $\chi^2$, or the Hellinger distance), the Bottleneck distance lacks a normalized proximity scale (e.g., {0,1}, where 0 indicates dissimilarity and 1 signifies equivalence). A potential solution is to define a scale-invariant Bottleneck distance for persistence diagrams.

\begin{definition}
Given the distance between two functions $f$ and $g$  with the $\mathcal{L}_\infty$ norm ($\leq \| f - g \|_\infty$), and the distance between their corresponding persistence diagrams, $D(f)$ and $D(g)$, measured using the Bottleneck distance $d_B$, the fundamental stability theorem in persistent homology \cite{Cohen-Steiner2005} bounds $d_B$ as:

\begin{equation}
    d_B(D(f), D(g)) \leq \| f - g \|_\infty
\end{equation}

where $\| f - g \|_\infty := \max(|f_i - g_i|)$ for $i \in \{1, \dots, N\}$ and $N$ is the number of elements in $f$ and $g$.
This definition implies that persistence diagrams are stable under small irregular perturbations and applies both to the Bottleneck distance and the Hausdorff distance. Another implication of this theorem is that the Bottleneck distance is bounded below by the Hausdorff distance and above by the supremum distance between the topological spaces $X$ and $Y$:

\begin{equation}
    d_H(D(f), D(g)) \leq d_B(D(f), D(g)) \leq \| f - g \|_\infty
\end{equation}

which means that the maximum difference between the distance functions of two geometric shapes equals the Hausdorff distance between them \cite{Buchet2016}.
\end{definition}

This definition  implies that if the metric used to measure the distance between $D$ and $D'$ is the Bottleneck distance, then, according to the fundamental stability theorem in persistent homology, this measure is relevant for comparing whether the topological spaces $X$ and $Y$ are congruent \cite{Cohen-Steiner2005}. Therefore, the Bottleneck distance is a direct measure that allows us to determine whether the topological signatures of two datasets are equal.

\begin{definition}
Given two finite compact spaces $(X, d_X)$ and $(Y, d_Y)$ with metrics $d_X$ and $d_Y$, respectively, we define the normalized Bottleneck distance as:

\begin{equation}
    d_{NB} = d_B \left( \frac{X}{\text{diam}(X)}, \frac{Y}{\text{diam}(Y)} \right)
\end{equation}

where $d_B$ is the Bottleneck distance according to definition 2.4.7.1. This pseudo-metric has the property of scale invariance:

\begin{equation}
    d_{NB}(aX, bY) = d_{NB}(X, Y) \quad \forall a, b > 0
\end{equation}
\end{definition}

The distance $d_{NB}$ is a pseudo-metric as it satisfies the properties of a metric except that it allows the distance between two different points to be zero. Consequently, $d_{NB}$ satisfies the following properties:

\begin{enumerate}
    \item Positivity: $d_{NB}(X, Y) \geq 0$
    \item Symmetry: $d_{NB}(X, Y) = d_{NB}(Y, X)$
    \item Triangle inequality: $d_{NB}(X, Z) \leq d_{NB}(X, Y) + d_{NB}(Y, Z)$
    \item Identity of indiscernibles: $d_{NB}(X, X) = d_{NB}(Y, Y) = 0$
\end{enumerate}

\begin{definition}
Given two subspaces $X$ and $Y$ of a larger metric space $Z$, $d_{NB}$ would be a pseudo-metric of $Z$ as there exist at least two points $X \neq Y$ such that $d_{NB}(X, Y) = 0$.

By the fundamental stability theorem in persistent homology,

\begin{equation}
    d_B(D(f), D(g)) \leq \| f - g \|_\infty
\end{equation}

the normalized Bottleneck distance $d_{NB}$ must satisfy:

\begin{equation}
    d_{NB} = d_B \left( \frac{X}{\text{diam}(X)}, \frac{Y}{\text{diam}(Y)} \right) \leq 1
\end{equation}
\end{definition}
Since $\| f - g \|_\infty = 1$, we can define a scale that limits the distance between two persistence diagrams within a range $0 \leq d_{NB} \leq 1$, where 1 indicates that the two diagrams are completely incongruent and 0 that they are entirely congruent.
\subsubsection{Introducing a new metric}
When evaluating synthetic data generated by GANs or other generative models, we need a principled approach to quantify the topological similarity between the original and generated data spaces. While the Bottleneck distance provides a measure of dissimilarity between persistence diagrams, it lacks standardization across different datasets, making comparative analysis difficult. Moreover, traditional metrics often fail to provide a clear interpretability threshold.

We propose a normalized metric derived from topological principles that quantifies the distance between two topological spaces in a bounded range $[0,1]$. The fundamental insight driving this approach is that randomly generated data, drawn from the same initial distribution used to train the GAN, should be topologically more distant from the original data than properly generated synthetic data. This property provides a natural baseline for normalization.

More formally, if we denote the original data space as $O$, synthetic data space as $S$, and random data space generated from the initial distribution as $A$, a well-performing generative model should satisfy $d_B(O,S) < d_B(O,A)$, where $d_B$ represents the Bottleneck distance between persistence diagrams. This relationship forms the basis for our normalized metric.

\begin{definition}[Bottleneck Metric]\label{def:main_theorem}
Given persistence diagrams $D(O)$, $D(S)$, and $D(A)$ corresponding to original $(O)$, synthetic $(S)$, and random $(A)$ spaces, where $O, S, A \subset \Omega$ and $S, A \in \mathbb{R}^n$, the normalized Bottleneck distance $d_{NB}(D(O), D(S))$ satisfies:

\begin{equation}
0 \leq d_B \left( \frac{D(O)}{diam(D(O))}, \frac{D(S)}{diam(D(S))} \right) \leq d_B \left( \frac{D(O)}{diam(D(O))}, \frac{D(A)}{diam(D(A))} \right)
\end{equation}

That is, the congruence between input data and generated synthetic data will always be greater than the congruence between input data and random samples, where in this case $Z$ and equivalent to the initial $\mathbb{P}_g$ have been used to train the GAN.

From the previous expression, we derive:

\begin{equation}
0 \leq \frac{d_B \left( \frac{D(O)}{diam(D(O))}, \frac{D(S)}{diam(D(S))} \right)}{d_B \left( \frac{D(O)}{diam(D(O))}, \frac{D(A)}{diam(D(A))} \right)} \leq 1
\end{equation}
\end{definition}

We define the Bottleneck metric $M_B$ as follows:
\begin{equation}
M_B =\frac{d_B \left( \frac{D(O)}{diam(D(O))}, \frac{D(S)}{diam(D(S))} \right)}{d_B \left( \frac{D(O)}{diam(D(O))}, \frac{D(A)}{diam(D(A))} \right)}
\end{equation}
where $0 \leq M_B \leq 1$. This metric $M_B$ satisfies the following conditions:
\begin{enumerate}
\item Positivity: $M_B(X,Y) \geq 0$
\item Symmetry: $M_B(X,Y) = M_B(Y,X)$
\item Triangle inequality: $M_B(X,Z) \leq M_B(X,Y) + M_B(Y,Z)$
\item Identity of indiscernibles: $M_B(X,X) = M_B(Y,Y) = 0$
\end{enumerate}

This normalized metric provides several advantages over the raw Bottleneck distance. First, it establishes a standardized range of $[0,1]$, facilitating intuitive interpretation: values closer to 0 indicate higher topological similarity between original and synthetic data, while values approaching 1 suggest that synthetic data is as topologically distant from the original data as random noise. Second, by incorporating the diameter normalization, the metric becomes scale-invariant, allowing for meaningful comparisons across datasets with different intrinsic scales. Third, by using the random data distribution as a reference point, we establish a contextually relevant upper bound that reflects the specific characteristics of the data domain.

However, this approach has some limitations. The quality of the normalized metric depends on how well the random distribution $A$ (random data space generated from the initial distribution) represents a true "null model" for comparison. Different choices for generating random data could yield different normalizations. Furthermore, the normalization by persistence diagram diameters assumes that the scale of topological features is relevant to the comparison, which may not hold in all applications.
\section{Experimental work}
\subsection{Materials and methods}
For basic statistical analyses, the \textit{Scipy} and \textit{Scikit-learn} libraries for Python are used. For calculating the Vietoris-Rips filtration we have used the \textit{Ripser} Python library \cite{Tralie2016}. WGAN architecture is shown in Algorithm 1.

\begin{algorithm}
\caption{Wasserstein GAN algorithm used in experiments. Adapted from \cite{Arjovsky2017}}
\begin{algorithmic}
\While{$\theta$ has not converged}
   \For{$t = 0,\ldots,n_d$}
       \State Sample $\{x^{(i)}\}_{i=1}^m \sim \mathbb{P}_r$, a batch of real data
       \State Sample $\{z^{(i)}\}_{i=1}^m \sim p(z)$, a batch of prior noise
       \State Compute discriminator gradient:
       \[
       g_w \leftarrow \nabla_w \left[\frac{1}{m} \sum_{i=1}^m f_w(x^{(i)}) - \frac{1}{m} \sum_{i=1}^m f_w(G_\theta(z^{(i)}))\right]
       \]
       \State Update discriminator weights: 
       \[
       w \leftarrow w - \alpha \cdot \text{RMSprop}(w, g_w)
       \]
       \State Clip weights: 
       \[
       w \leftarrow \text{clip}(w, -c, c)
       \]
   \EndFor
   \State Sample $\{z^{(i)}\}_{i=1}^m \sim p(z)$, a batch of prior noise
   \State Compute generator gradient:
   \[
   g_\theta \leftarrow \nabla_\theta \left[ \frac{1}{m} \sum_{i=1}^m f_w(G_\theta(z^{(i)})) \right]
   \]
   \State Update generator weights:
   \[
   \theta \leftarrow \theta - \alpha \cdot \text{RMSprop}(\theta, g_\theta)
   \]
\EndWhile
\end{algorithmic}
\end{algorithm}

\subsection{Datasets use in experiments}
We have used four different datasets in our experiments. The first dataset is named e-scotter, a dataset for marketing analysis that we have synthetically created. This data shows the relation between e-scooters prices and some marketing features. The data has a similar number of variables and observations, n=11 observations and m=9 variables. Data has both continuous and categorical variables (Table \ref{tab:escooter-competitors}). 

\begin{table}[htbp]
\centering
\caption{E-Scooter Competitor Analysis}
\label{tab:escooter-competitors}
\footnotesize 
\setlength{\tabcolsep}{3pt} 
\begin{tabularx}{\textwidth}{lrrrrrrcXr}
\toprule
Competitor & Price & Years & Feature & Feature & Feature & Feature & Marketing & Brand & Market \\
 & (\$) & in market & 1 & 2 & 3 & 4 & Effort & Awareness (\%) & share (\%) \\
\midrule
Competitor 1 & 3400.00 & 3 & 40 & 60 & 50 & 50 & HIGH & 78.00 & 9.56 \\
Competitor 2 & 3500.00 & 4 & 50 & 70 & 50 & 50 & HIGH & 80.00 & 20.23 \\
Competitor 3 & 3550.00 & 5 & 90 & 50 & 55 & 70 & MEDIUM & 93.00 & 25.21 \\
Competitor 4 & 3500.00 & 4 & 50 & 70 & 60 & 50 & MEDIUM & 75.00 & 14.60 \\
Competitor 5 & 5000.00 & 3 & 45 & 50 & 45 & 50 & HIGH & 78.00 & 7.80 \\
Competitor 6 & 3000.00 & 2 & 35 & 40 & 55 & 45 & HIGH & 55.00 & 6.70 \\
Competitor 7 & 2000.00 & 3 & 15 & 45 & 35 & 40 & LOW & 61.00 & 4.90 \\
Competitor 8 & 2350.00 & 4 & 30 & 30 & 40 & 30 & LOW & 68.00 & 5.00 \\
Competitor 9 & 2850.00 & 2 & 40 & 25 & 20 & 20 & LOW & 40.00 & 4.00 \\
Competitor 10 & 3450.00 & 3 & 20 & 35 & 40 & 30 & MEDIUM & 55.00 & 3.00 \\
Competitor 11 & 1900.00 & 1 & 15 & 30 & 15 & 20 & LOW & 30.00 & 1.00 \\
\bottomrule
\end{tabularx}
\end{table}

The second dataset represents a model of a SOFC (Solid Oxide Fuel Cell) hydrogen battery. The data has been synthetically created from a physical model to simulate the battery intensity as a function of certain input parameters as described in \cite{Lakshmi2013}. The dataset has 30 rows and 15 columns. This dataset is particularly useful as we can directly compare the general synthetic data with the data obtained from the mathematical model that we will describe now. The limitation of this dataset is that it only contains continuous variables and does not include any categorical variable. The available columns in SOFC dataset are: Temperature, $I_{fc}$ Current, $q_{H2}$ (Hydrogen input flow), $q_{O2}$ (Oxygen input flow), $P_{pH2}$(Partial pressure of Hydrogen), $P_{pO2}$ (Partial pressure of Oxygen), $P_{pH2O}$ (partial pressure of water vapor), Nernst Reversible Voltage, Activation polarization, Concentration polarization, Ohmic polarization, Cell voltage (calculated according \cite{Lakshmi2013}), Cell Power Observed voltage (experimentally measured voltage), and difference between experimentally measured voltage and theoretical voltage. The cell voltage has been calculated using the following equation \cite{Lakshmi2013}:

\begin{equation}
V_{dc} = E_{nernst} - \eta_{act} - \eta_{conc} - \eta_{ohmic}
\end{equation}

The third dataset is called Balls and describes the well-known flotation effect of different balls when falling from a certain height and how it influences the time it takes for the ball to reach the ground \cite{deSilva2020}. This is a very well-known experiment that is frequently used in secondary schools due to its visual appeal.

The last dataset has the results from the last nine USA presidential elections collected by MIT Election Data and Science Lab \cite{MIT2021}. Data has fifty-two columns (USA states) and an additional column with a categorical variable defining the winner (republicans/democrats). However, it has only nine rows.

\subsection{Experimental results}

Table \ref{tab:global-metrics} presents the results of global data metrics. Propensity scores range from $(0, +\infty)$ indicating a high similarity for low values. Large cluster analysis values reveal membership disparities between original and synthetic data, while negative values suggest high similarity. The SOFC dataset shows stronger similarity according to both pMSE and MMD measures, though both datasets have values close to zero (excluding cluster analysis), indicating statistical equivalence between synthetic and input data. The variation in metrics—with pMSE suggesting SOFC generated data more closely matches its original than e-scotter— highlights why synthetic data validation requires multiple complementary metrics rather than relying on any single measure to reach valid conclusions.

\begin{table}[htbp]
\centering
\caption{Global metrics comparison for SOFC and e-scooter}
\label{tab:global-metrics}
\begin{tabular}{lcc}
\toprule
Dataset$\rightarrow$ & SOFC & e-scotter \\
\midrule
Number of rows & 30 & 11 \\
Number of columns & 13 & 9 \\
Ratio rows/columns & 2.3 & 1.2 \\
\midrule
\multicolumn{3}{c}{Global Metrics} \\
\midrule
Propensity score* pMSE & 0.020 & 0.021 \\
Cluster analysis measure*, c & -0.38 & -0.28 \\
MMD (Maximum Mean Discrepancy)* & 0.0367 & 0.100 \\
\bottomrule
\multicolumn{3}{l}{* Number of synthetically generated samples = 10 $\times$ $n$} \\
\multicolumn{3}{l}{$n$: Number of data samples} \\
\end{tabular}
\end{table}

Our method uses univariate non-parametric tests to evaluate topological signatures of original and synthetic data in homology group $H_0$. We first compute persistence barcode distributions for both datasets and then apply statistical tests to these distributions. Additionally, we calculate the Bottleneck distance between persistence diagrams in homology group $H_0$ to measure how close their topological signatures are. While this distance cannot compare similarity across different datasets, it effectively evaluates different augmentation ratios within the same dataset. Tables \ref{tab:sofc-metrics} and \ref{tab:escotter-metrics} summarize our findings on global metrics, Bottleneck distances, and non-parametric test results.

We have first computed the persistence barcodes distributions of original and synthetic data and then applied univariate non-parametric statistical tests. For these tests, we have used solely $H_0$ dimension. We have computed the Bottleneck distance between original and synthetic data persistence diagrams. Tables \ref{tab:sofc-metrics} and \ref{tab:escotter-metrics} show the results of comparing global metrics, distances (Bottleneck) and non-parametric statistical tests.

\begin{table}[htbp]
\centering
\caption{Metrics comparison for SOFC dataset with different number of samples}
\label{tab:sofc-metrics}
\begin{tabular}{lcccc}
\toprule
Augmentation rate $\rightarrow$ & 5 $\times$ n & 10 $\times$ n & 20 $\times$ n & 50 $\times$ n \\
\midrule
Propensity score* pMSE & 0.03 & 0.02 & 0.05 & 0.08 \\
Cluster analysis measure*, c & 0.37 & -0.38 & -0.57 & -0.59 \\
MMD (Maximum Mean Discrepancy)* & 0.040 & 0.0367 & 0.0350 & 0.034 \\
\midrule
Bottleneck distance, $d_B$ & 31.45 & 16.31 & 11.06 & 16.59 \\
\midrule
Kolmogorov-Smirnov (p-value)* & 0.4175 & 0.4175 & 0.4175 & 0.0524 \\
Mann-Whitney U test (p-value)* & 0.6193 & 0.2687 & 0.4917 & 0.0351 \\
Chi-square test (p-value)* & 0.1848 & 0.4454 & 0.4454 & 0.377 \\
\bottomrule
\multicolumn{5}{l}{* Test performed in homology group $H_0$} \\
\multicolumn{5}{l}{* Significance level $\alpha = 0.05$} \\
\multicolumn{5}{l}{$n$: Number of data samples} \\
\end{tabular}
\end{table}

Global metrics demonstrate high similarity between original and synthetic data across both datasets, with consistently low pMSE and MMD values. However, the SOFC dataset shows significantly elevated pMSE at high augmentation rates, suggesting excessive augmentation may introduce substantial noise into the synthetic dataset.

For the SOFC dataset, Tables \ref{tab:sofc-metrics} and \ref{tab:escotter-metrics} show non-parametric tests confirm original and synthetic data arise from identical distributions (95\% confidence) for all augmentation rates except 50×. Conversely, the e-scotter dataset tests indicate different distributions regardless of augmentation rate. The degradation at high augmentation rates in SOFC but not in e-scotter suggests absolute sample count rather than rate may be critical—1500 synthetic samples for SOFC versus 450 for e-scotter at 50× augmentation potentially introduce noise beyond statistical validity thresholds.

\begin{table}[htbp]
\centering
\caption{Metrics comparison for e-scotter dataset with different number of samples}
\label{tab:escotter-metrics}
\begin{tabular}{lcccc}
\toprule
Augmentation rate $\rightarrow$ & 5 $\times$ n & 10 $\times$ n & 20 $\times$ n & 50 $\times$ n \\
\midrule
Propensity score* pMSE & 0.1091 & 0.021 & 0.023 & 0.11 \\
Cluster analysis measure*, c & -0.14 & -0.28 & -0.15 & 1.18 \\
MMD (Maximum Mean Discrepancy)* & 0.1091 & 0.100 & 0.095 & 0.0927 \\
\midrule
Bottleneck distance, $d_B$ & 725.53 & 725.53 & 725.53 & 725.53 \\
\midrule
Kolmogorov-Smirnov (p-value)* & 0.0021 & 0.0002 & 0.0002 & 0.0002 \\
Mann-Whitney U test (p-value)* & 0.0018 & 0.0002 & 0.0007 & 0.0002 \\
Chi-square test (p-value)* & 0.00725 & 0.0186 & 0.0671 & 0.0103 \\
\bottomrule
\multicolumn{5}{l}{* Test performed in homology group $H_0$} \\
\multicolumn{5}{l}{* Significance level $\alpha = 0.05$} \\
\multicolumn{5}{l}{$n$: Number of data samples} \\
\end{tabular}
\end{table}

\begin{table}[htbp]
\centering
\caption{Metrics comparison several Low Sample Size datasets}
\label{tab:all-datasets}
\begin{tabular}{lcccc}
\toprule
Dataset$\rightarrow$ & SOFC & e-scotter & Balls & USA elections \\
\midrule
Number of rows original & 30 & 11 & 9 & 9 \\
Number of columns & 13 & 9 & 7 & 56 \\
Ratio rows/columns original & 2.3 & 1.2 & 1.3 & 0.16 \\
\midrule
\multicolumn{5}{c}{Global Metrics} \\
\midrule
Propensity score, pMSE & 0.0367 & 0.100 & 0.0356 & 0.0507 \\
Cluster analysis measure, c & -0.38 & -0.28 & 1.33 & -0.25 \\
MMD (Maximum Mean Discrepancy) & 0.020 & 0.021 & 0.1222 & 0.1229 \\
\midrule
\multicolumn{5}{c}{Topological data analysis} \\
\midrule
Bottleneck distance, $d_B$ & 16.31 & 725.53 & 297000 & 6.07 \\
\midrule
Kolmogorov-Smirnov (p-value) * & 0.4175 & 0.0002 & 0.0002 & 0.0123 \\
Mann-Whitney U test (p-value) * & 0.2687 & 0.0002 & 0.0004 & 0.0055 \\
Chi-square test (p-value) * & 0.4454 & 0.0186 & 0.0246 & 0.0478 \\
\bottomrule
\multicolumn{5}{l}{* Test performed in homology group $H_0$} \\
\multicolumn{5}{l}{* Significance level $\alpha = 0.05$} \\
\multicolumn{5}{l}{$n$: Number of data samples} \\
\end{tabular}
\end{table}

Table \ref{tab:all-datasets} illustrates that there is no direct correlation between the global metrics and non-parametric statistical tests in homology group $H_0$ across all four datasets. For example, the propensity score and MMD values for SOFC, e-scotter, and Balls datasets are similarly close to zero, indicating that original and synthetic data appear statistically equivalent using these global metrics. However, the USA elections dataset shows higher MMD values despite having low propensity scores, revealing inconsistencies in global metric assessments. 

The Bottleneck distances exhibit significant variations across datasets (from 6.07 for USA elections to 297000 for Balls), further demonstrating the challenge of direct cross-dataset comparisons using topological measures. When examining non-parametric statistical tests, only the SOFC dataset shows p-values consistently above the 0.05 significance threshold (Kolmogorov-Smirnov: 0.4175, Mann-Whitney: 0.2687, Chi-square: 0.4454), allowing us to conclude that only in this case were original and synthetic data likely drawn from the same distribution at the 95\% confidence level.

The Balls dataset shows particularly interesting results, with the highest cluster analysis measure (1.33) suggesting significant structural differences between original and synthetic data, confirmed by extremely low p-values in non-parametric tests and the largest Bottleneck distance. Similarly, while USA elections data has the lowest Bottleneck distance (6.07), its non-parametric test p-values are still below or just at the significance threshold, suggesting topological similarity doesn't necessarily translate to distribution equivalence. 

These findings underscore the importance of using multiple complementary assessment methods when validating synthetic data quality, especially for low sample size datasets. We are going to use the proposed metric $M_B$ to get an additional comparison approach using topological data spaces distance.

\begin{table}[htbp]
\centering
\caption{Comprehensive evaluation metrics for SOFC dataset}
\label{tab:sofc2-metrics}
\small 
\setlength{\tabcolsep}{5pt} 
\begin{tabular}{lccccccc}
\toprule
\multicolumn{8}{c}{\textbf{Global Tests}} \\
\midrule
Metric & R1 & R2 & R3 & R4 & R5 & Mean & SD \\
\midrule
MMD & 0.0366 & 0.0367 & 0.0367 & 0.0366 & 0.0367 & 0.0367 & 0.0001 \\
Cluster analysis & -0.53 & 0.27 & -0.53 & 0.30 & -0.60 & -0.218 & 0.460 \\
pMSE & 0.0467 & 0.1585 & 0.1327 & 0.1512 & 0.0288 & 0.104 & 0.061 \\
\midrule
\multicolumn{8}{c}{\textbf{Topological Data Analysis}} \\
\midrule
$d_B$ & 19.57 & 11.46 & 11.34 & 11.01 & 12.45 & 13.17 & 3.62 \\
K-S test & 0.0524 & 0.1678 & 0.1678 & 0.1678 & 0.1678 & 0.145 & 0.052 \\
M-W test & 0.0348 & 0.0723 & 0.1817 & 0.0723 & 0.0663 & 0.086 & 0.056 \\
Chi-sq. test & 0.3779 & 0.3142 & 0.2794 & 0.3142 & 0.3142 & 0.320 & 0.036 \\
$M_B$ (\%) & 75.42 & 80.22 & 118.48 & 60.06 & 64.45 & 79.72 & 35.25 \\
\bottomrule
\end{tabular}
\end{table}

\begin{table}[H]
\centering
\caption{Comprehensive evaluation metrics for Balls dataset}
\label{tab:balls-metrics}
\small
\setlength{\tabcolsep}{5pt}
\begin{tabular}{lccccccc}
\toprule
\multicolumn{8}{c}{\textbf{Global Tests}} \\
\midrule
Metric & R1 & R2 & R3 & R4 & R5 & Mean & SD \\
\midrule
MMD & 0.1222 & 0.1217 & 0.1222 & 0.1222 & 0.1222 & 0.1221 & 0.0002 \\
Cluster analysis & -0.39 & 1.33 & 0.08 & -1.41 & -1.41 & -0.3600 & 1.1463 \\
pMSE & 0.0222 & 0.0210 & 0.0049 & 0.0059 & 0.0031 & 0.0114 & 0.0094 \\
\midrule
\multicolumn{8}{c}{\textbf{Topological Data Analysis}} \\
\midrule
$d_B$ & 297000 & 297000 & 297000 & 297000 & 297000 & 297000 & 0.0 \\
K-S test & 0.0002 & 0.0002 & 0.0002 & 0.0002 & 0.0002 & 0.0002 & 0.0000 \\
M-W test & 0.0001 & 0.0002 & 0.0002 & 0.0002 & 0.0002 & 0.0002 & 0.0000 \\
Chi-sq. test & 0.0196 & 0.0028 & 0.0028 & 0.0028 & 0.0028 & 0.0060 & 0.0075 \\
$M_B$ (\%) & 100.22 & 84.34 & 60.13 & 75.08 & 20.31 & 68.02 & 30.38 \\
\bottomrule
\end{tabular}
\end{table}

\begin{table}[H]
\centering
\caption{Comprehensive evaluation metrics for USA elections dataset}
\label{tab:usa-metrics}
\small
\setlength{\tabcolsep}{5pt}
\begin{tabular}{lccccccc}
\toprule
\multicolumn{8}{c}{\textbf{Global Tests}} \\
\midrule
Metric & R1 & R2 & R3 & R4 & R5 & Mean & SD \\
\midrule
MMD & 0.1225 & 0.1254 & 0.1225 & 0.1238 & 0.1287 & 0.1246 & 0.0026 \\
Cluster analysis & 0.50 & -0.16 & -0.25 & -0.39 & 0.56 & 0.0520 & 0.4445 \\
pMSE & 0.0415 & 0.0132 & 0.0754 & 0.0233 & 0.0218 & 0.0350 & 0.0248 \\
\midrule
\multicolumn{8}{c}{\textbf{Topological Data Analysis}} \\
\midrule
$d_B$ & 6.07 & 6.07 & 6.07 & 6.07 & 6.07 & 6.07 & 0.0 \\
K-S test & 0.0123 & 0.0017 & 0.0123 & 0.0123 & 0.0123 & 0.0102 & 0.0046 \\
M-W test & 0.0080 & 0.0021 & 0.0080 & 0.0104 & 0.0075 & 0.0071 & 0.0032 \\
Chi-sq. test & 0.0880 & 0.1432 & 0.4780 & 0.2603 & 0.0478 & 0.2835 & 0.1997 \\
$M_B$ (\%) & 51.24 & 57.48 & 58.95 & 87.67 & 45.48 & 60.16 & 16.36 \\
\bottomrule
\end{tabular}
\end{table}

Our comprehensive analysis results (we have repeated each experiment five times) of synthetic data evaluation across four datasets are shown in tables \ref{tab:sofc2-metrics}, \ref{tab:balls-metrics}, and \ref{tab:usa-metrics}  critical limitations in both previous metrics and our proposed Bottleneck metric $M_B$. While global metrics (pMSE, MMD) show reasonable consistency across experimental runs, they frequently contradict topological measures—notably in the Balls dataset where low pMSE (0.0114) suggests similarity despite extreme Bottleneck distance and statistically significant differences (p-values near 0.0002). The $M_B$ metric, though theoretically bounded [0,1], demonstrates concerning variability (standard deviations: SOFC 35.25\%, Balls 30.38\%, USA 16.36\%) and occasionally exceeds theoretical bounds (maximum 118.48\% in SOFC), suggesting fundamental instability in normalization approaches.

\section{Discussion and future work}
In this paper, we investigated the challenge of generating and validating synthetic tabular data from very small sample sizes. We proposed a method combining topology and robust statistics for hypothesis testing to assess synthetic data quality, introducing a novel normalized Bottleneck distance based metric ($M_B$). Our comprehensive evaluation across four low-sample datasets shows several important findings.

Our results demonstrate that traditional global metrics (pMSE, MMD) frequently contradict topological measures, raising critical questions about their reliability for low-sample data. For instance, the Balls dataset exhibited reasonably low pMSE values (0.0114) despite an extreme Bottleneck distance (297,000.19) and statistically significant differences in distribution. This contradiction exposes a fundamental limitation in current evaluation approaches that might mislead researchers about synthetic data quality.
The proposed $M_B$ metric successfully identifies cases where global metrics fail to capture topological dissimilarities, providing valuable complementary information. However, its high variability across experimental runs (standard deviations: SOFC 35.25\%, Balls 30.38\%, USA 16.36\%) and occasional values exceeding theoretical bounds indicate inherent instability in the normalization approach. This instability appears particularly pronounced in extremely low-sample datasets, suggesting sample size sensitivity that must be addressed in future work.

Dataset-specific results further highlight how topology-based methods can reveal structural differences invisible to traditional metrics. The SOFC dataset demonstrated reasonable consistency in global metrics but showed marked variability in topological measures. Conversely, the USA elections dataset maintained the lowest Bottleneck distance (6.07) while still failing distributional equivalence tests, revealing that topological similarity doesn't guarantee distributional equivalence.
The primary contribution of this research lies in demonstrating the necessity of multiple complementary assessment methods for synthetic data validation. Neither global metrics nor topological measures alone provide sufficient information, especially for low-sample datasets. Additionally, our work establishes a theoretical foundation for normalizing topological distances that, despite current limitations, offers a promising direction for future research.
Future work should focus on improving $M_B$ stability through new normalization techniques and establishing theoretical bounds on its expected variability as a function of sample size. Developing robust statistical methods specifically designed for low-sample persistence diagrams would also significantly advance this field. Integration of these topological measures into GAN training processes could potentially guide generators toward preserving basic topological features, potentially improving synthetic data quality for small datasets.

In conclusion, while our proposed method shows promise in complementing traditional evaluation approaches, its current limitations emphasize the inherent difficulty of synthetic data generation and validation with very low samples. Nevertheless, this research establishes an important framework for combining topological data analysis with statistical methods that can be refined to address these challenges in future work.


\begin{thebibliography}{99}

\bibitem{Arjovsky2017} Arjovsky, M., Chintala, S., \& Bottou, L. (2017). Wasserstein Generative Adversarial Networks. International Conference on Machine Learning, 214--223.

\bibitem{Barnes2021} Barnes, D., Polanco, L., \& Perea, J. A. (2021). A Comparative Study of Machine Learning Methods for Persistence Diagrams. Frontiers in Artificial Intelligence, 4, 91.

\bibitem{Benaim2020} Benaim, A. R., Almog, R., Gorelik, Y., Hochberg, I., Nassar, L., Mashiach, T., Khamaisi, M., Lurie, Y., Azzam, Z. S., Khoury, J., Kurnik, D., \& Beyar, R. (2020). Analyzing Medical Research Results Based on Synthetic Data and Their Relation to Real Data Results: Systematic Comparison From Five Observational Studies. JMIR Med Inform 2020;8(2):E16492.

\bibitem{Beyer1998} Beyer, K., Goldstein, J., Ramakrishnan, R., \& Shaft, U. (1998). When is "nearest neighbor" meaningful? Lecture Notes in Computer Science (Including Subseries Lecture Notes in Artificial Intelligence and Lecture Notes in Bioinformatics), 1540, 217--235.

\bibitem{Blumberg2014} Blumberg, A. J., Gal, I., Mandell, M. A., \& Pancia, M. (2014). Robust Statistics, Hypothesis Testing, and Confidence Intervals for Persistent Homology on Metric Measure Spaces. Foundations of Computational Mathematics 2014 14:4, 14(4), 745--789.

\bibitem{Borgwardt2006} Borgwardt, K. M., Gretton, A., Rasch, M. J., Kriegel, H.-P., Schö Lkopf, B., \& Smola, A. J. (2006). Integrating structured biological data by Kernel Maximum Mean Discrepancy. Bioinformatics, 22(14), 49--57.
\bibitem{Buchet2016} Buchet, M., Chazal, F., Oudot, S. Y., \& Sheehy, D. R.
(2016). Efficient and robust persistent homology for measures. Computational Geometry, 58, 70–96.

\bibitem{Carlsson2009} Carlsson, G. (2009). Topology and data. Bull. Amer. Math. Soc., 46(2), 255--308.

\bibitem{Carlsson2004} Carlsson, G., Zomorodian, A., Collins, A., \& Guibas, L. (2004). Persistence barcodes for shapes. ACM International Conference Proceeding Series, 71, 124--135.

\bibitem{Chaudhari2020} Chaudhari, P., Agrawal, H., \& Kotecha, K. (2020). Data augmentation using MG-GAN for improved cancer classification on gene expression data. Soft Computing, 24(15), 11381--11391.

\bibitem{Cheng2019} Cheng, S., Dong, L., Yu, Z., Hao, T., Liu, R., \& Chen, G. (2019). Improving the Generation Quality of GAN via Self-modified Discriminator. 2019 IEEE International Conference on Systems, Man and Cybernetics (SMC), 1242--1247.

\bibitem{Chundawat2022} Chundawat, V. S., Tarun, A. K., Mandal, M., Lahoti, M., \& Narang, P. (2022). TabSynDex: A Universal Metric for Robust Evaluation of Synthetic Tabular Data. ArXiv.Org.

\bibitem{Cohen-Steiner2005} Cohen-Steiner, D., Edelsbrunner, H., \& Harer, J. (2005). Stability of persistence diagrams. Proceedings of the Annual Symposium on Computational Geometry, 263--271.

\bibitem{deSilva2020} de Silva, B. M., Higdon, D. M., Brunton, S. L., \& Kutz, J. N. (2020). Discovery of Physics From Data: Universal Laws and Discrepancies. Frontiers in Artificial Intelligence, 3, 25.

\bibitem{Elman1993} Elman, J. L. (1993). Learning and development in neural networks: the importance of starting small. Cognition, 48(1), 71--99.

\bibitem{Eno2008} Eno, J., \& Thompson, C. W. (2008). Generating synthetic data to match data mining patterns. IEEE Internet Computing, 12(3), 78--82.

\bibitem{Goncalves2020} Goncalves, A., Ray, P., Soper, B., Stevens, J., Coyle, L., \& Sales, A. P. (2020). Generation and evaluation of synthetic patient data. BMC Medical Research Methodology, 20(1), 1--40.

\bibitem{Goodfellow2014} Goodfellow, I. J. (2014). ON DISTINGUISHABILITY CRITERIA FOR ESTIMATING GENERATIVE MODELS. ArXiv Preprint.

\bibitem{Goodfellow2014b} Goodfellow, I. J., Pouget-Abadie, J., Mirza, M., Xu, B., Warde-Farley, D., Ozair, S., Courville, A., \& Bengio, Y. (2014). Generative Adversarial Nets. Advances in Neural Information Processing Systems, 27.

\bibitem{Gretton2006} Gretton, A., Borgwardt, K. M., Rasch, M., Schölkopf, B., \& Smola, A. J. (2006). A kernel method for the two-sample-problem. Advances in Neural Information Processing Systems, 19.

\bibitem{Gretton2009} Gretton, A., Fukumizu, K., Harchaoui, Z., \& Sriperumbudur, B. K. (2009). A Fast, Consistent Kernel Two-Sample Test. Advances in Neural Information Processing Systems, 22.

\bibitem{Hou2022} Hou, X., Shen, Y., Sun, H., Zhao, B., \& Zhou, H. (2022). DP-CGAN: Differentially Private Synthetic Data and Label Generation. In IEEE Conference on Computer Vision and Pattern Recognition Workshops (pp. 2666--2676).

\bibitem{Karras2018} Karras, T., Aila, T., Laine, S., \& Lehtinen, J. (2018). Progressive Growing of GANs for Improved Quality, Stability, and Variation. International Conference on Learning Representations.

\bibitem{Lakshmi2013} Lakshmi, T., \& Geethanjali, P. (2013). Mathematical modelling of solid oxide fuel cell using Matlab/Simulink. In IEEE (Ed.), 2013 Annual International Conference on Emerging Research Areas and 2013 International Conference on Microelectronics, Communications and Renewable Energy (pp. 1--5).

\bibitem{Lakshmi2013b} Lakshmi, T. V. V. S., Geethanjali, P., \& Krishna Prasad, S. (2013). Mathematical modelling of solid oxide fuel cell using Matlab/Simulink. 2013 Annual International Conference on Emerging Research Areas and 2013 International Conference on Microelectronics, Communications and Renewable Energy.

\bibitem{Lin2015} Lin, J. J., Chang, C. H., \& Pal, N. (2015). A Revisit to Contingency Table and Tests of Independence: Bootstrap is Preferred to Chi-Square Approximations as Well as Fisher's Exact Test. Journal of Biopharmaceutical Statistics, 25(3), 438--458.

\bibitem{Liu2019} Liu, Y., Zhou, Y., Liu, X., Dong, F., Wang, C., \& Wang, Z. (2019). Wasserstein GAN-Based Small-Sample Augmentation for New-Generation Artificial Intelligence: A Case Study of Cancer-Staging Data in Biology. Engineering, 5(1), 156--163.

\bibitem{Massey1951} Massey, F. J. (1951). The Kolmogorov-Smirnov Test for Goodness of Fit. Journal of the American Statistical Association, 46(253), 68--78.

\bibitem{Mazumdar2020} Mazumdar, E., Ratliff, L. J., \& Sastry, S. S. (2020). On Gradient-Based Learning in Continuous Games. SIAM Journal on Mathematics of Data Science, 2(1), 103--131.

\bibitem{McKnight2010} McKnight, P. E., \& Najab, J. (2010). Mann-Whitney U Test. The Corsini Encyclopedia of Psychology, 1--1.

\bibitem{MIT2021} MIT Election Data and Science Lab. (2021). U.S. President 1976–2020. In Harvard Dataverse (Vol. 6). Harvard Dataverse. 

\bibitem{Nowozin2016} Nowozin, S., Cseke, B., \& Tomioka, R. (2016). f-gan: Training generative neural samplers using variational divergence minimization. Advances in Neural Information Processing Systems, 29.

\bibitem{Rashid2019} Rashid, S., \& Louis, S. J. (2019). Evolutionary approaches to generating synthetic tabular data. 2019 18th IEEE International Conference On Machine Learning And Applications (ICMLA), 1002--1009.

\bibitem{Rosenbaum1983} Rosenbaum, P. R., \& Rubin, D. B. (1983). The central role of the propensity score in observational studies for causal effects. Biometrika, 70(1), 41--55.

\bibitem{Snoke2018} Snoke, J., Raab, G. M., Nowok, B., Dibben, C., \& Slavkovic, A. (2018). General and specific utility measures for synthetic data. Journal of the Royal Statistical Society. Series A: Statistics in Society, 181(3), 663--688.

\bibitem{Snoke2018a} Snoke, J., \& Slavković, A. (2018). pMSE mechanism: Differentially private synthetic data with maximal distributional similarity. Lecture Notes in Computer Science (Including Subseries Lecture Notes in Artificial Intelligence and Lecture Notes in Bioinformatics), 11126 LNCS, 138--159.

\bibitem{Sutherland2016} Sutherland, D. J., Tung, H.-Y., Strathmann, H., De, S., Ramdas, A., Smola, A., \& Gretton, A. (2016). GENERATIVE MODELS AND MODEL CRITICISM VIA OPTIMIZED MAXIMUM MEAN DISCREPANCY. ArXiv Preprint ArXiv:1611.04488.

\bibitem{Sutherland2017} Sutherland, D. J., Tung, H.-Y., Strathmann, H., De, S., Ramdas, A., Smola, A., \& Gretton, A. (2017). GENERATIVE MODELS AND MODEL CRITICISM VIA OPTIMIZED MAXIMUM MEAN DISCREPANCY. ICLR 2017.

\bibitem{Theis2016} Theis, L., van den Oord, A., \& Bethge, M. (2016). A NOTE ON THE EVALUATION OF GENERATIVE MODELS. International Conference on Learning Representations.

\bibitem{Tralie2016} Tralie, C., Saul, N., \& Bar-On, R. (2018). Ripser. py: A lean persistent homology library for python. Journal of Open Source Software, 3(29), 925.

\bibitem{Virtanen2020} Virtanen, P., Gommers, R., Oliphant, T. E., Haberland, M., Reddy, T., Cournapeau, D., Burovski, E., Peterson, P., Weckesser, W., Bright, J., van der Walt, S. J., Brett, M., Wilson, J., Jarrod Millman, K., Mayorov, N., J Nelson, A. R., Jones, E., Kern, R., Larson, E., … van Mulbregt, P. (2020). SciPy 1.0: fundamental algorithms for scientific computing in Python. Nature Methods, 17, 261--272.

\bibitem{Vlontzos2021} Vlontzos, A., Cao, Y., Schmidtke, L., Kainz, B., \& Monod, A. (2021). Topological data analysis of database representations for information retrieval. CoRR abs/2104.01672, 2104.

\bibitem{Woo2009} Woo, M.-J., Reiter, J. P., Oganian, A., \& Karr, A. F. (2009a). Global Measures of Data Utility for Microdata Masked for Disclosure Limitation. Journal of Privacy and Confidentiality, 1(1), 111--124.


\bibitem{Xu2019} Xu, L., Skoularidou, M., Cuesta-Infante, A., \& Veeramachaneni, K. (2019). Modeling Tabular data using Conditional GAN. Advances in Neural Information Processing Systems, 32.

\bibitem{Xu2018} Xu, Q., Huang, G., Yuan, Y., Guo, C., Sun, Y., Wu, F., \& Weinberger, K. Q. (2018). An empirical study on evaluation metrics of generative adversarial networks. ArXiv Preprint.

\end{thebibliography}
\end{document}